\documentclass[a4paper,conference]{IEEEtran}
\IEEEoverridecommandlockouts
\usepackage{cite}
\usepackage{amsmath,amssymb,amsfonts}
\usepackage{algorithmic}
\usepackage{textcomp}
\usepackage{xcolor}

\usepackage{graphicx}
\usepackage{amsmath,amssymb} 
\usepackage{cuted}

\usepackage{cite}

\usepackage{times}
\usepackage{epsfig}
\usepackage{algorithmic}
\usepackage{xcolor}
\usepackage{multirow}
\usepackage{caption}
\usepackage{subcaption}
\usepackage{booktabs}
\usepackage{tabularx}
\usepackage{tikz}
\usetikzlibrary{shadows}
\usepackage{adjustbox}
\usepackage{fancybox}

\newcommand{\etal}{\textit{et al.}}
\newcommand{\ie}{\textit{i.e.}}

\ifCLASSINFOpdf
\else
\fi

\def\BibTeX{{\rm B\kern-.05em{\sc i\kern-.025em b}\kern-.08em
    T\kern-.1667em\lower.7ex\hbox{E}\kern-.125emX}}
\begin{document}
%
\title{Multi-Domain Image-to-Image Translation with Adaptive Inference Graph}



%

\author{\IEEEauthorblockN{The-Phuc Nguyen}
\IEEEauthorblockA{\textit{Department of Computer Vision}\\
\textit{Vingroup Big Data Institute}\\
Hanoi, Vietnam\\
v.phucnt22@vinbdi.org}
\and
\IEEEauthorblockN{St\'{e}phane Lathuili\`{e}re}
\IEEEauthorblockA{\textit{LTCI, T\'{e}l\'{e}com Paris, }\\
\textit{Institut Polytechnique de Paris}\\
Palaiseau, France \\
stephane.lathuiliere@telecom-paris.fr}
\and
\IEEEauthorblockN{Elisa Ricci}
\IEEEauthorblockA{\textit{FBK} and \textit{University of Trento} \\
Trento, Italy \\
e.ricci@unitn.it}
}


\maketitle

\begin{abstract}
In this work, we address the problem of multi-domain image-to-image translation with particular attention paid to computational cost. In particular, current state of the art models require a large and deep model in order to handle the visual diversity of multiple domains. In a context of limited computational resources, increasing the network size may not be possible. Therefore, we propose to increase the network capacity by using an adaptive graph structure. At inference time, the network estimates its own graph by selecting specific sub-networks. Sub-network selection is implemented using Gumbel-Softmax in order to allow end-to-end training. This approach leads to an adjustable increase in number of parameters while preserving an almost constant computational cost. Our evaluation on two publicly available datasets of facial and painting images shows that our adaptive strategy generates better images with fewer artifacts than literature methods.
\end{abstract}


%
\IEEEpeerreviewmaketitle

\section{Introduction}
\label{sec:intro}

The task of Image-to-image translation consists in learning a mapping between two different visual domains \cite{Isola_2017_CVPR,zhu2017unpaired}. For instance, daylight images can be changed to night images or the age of a person in a picture can be changed while preserving identity and pose \cite{Geng20193DGF}. Image-to-image translation has received a growing interest in the computer vision community since many problems, such as style transfer~\cite{huang2017arbitrary}, photo editing~\cite{Geng20193DGF}, image segmentation~\cite{couprie2018joint} or depth estimation~\cite{pilzer2019refine}, can be addressed following an image-to-image formulation. 

Image-to-image translation has been also addressed in the multi-domain setting where images are translated between multiple visual domains. This problem can be addressed by considering either multiple models, as in ComboGAN \cite{anoosheh2018combogan} or with a single model \cite{choi2018stargan}. The major difficulty of this problem lies in the high complexity of the models when the number of domains is large. In particular, when the number of domain increases, training generator networks for every pair of domains would lead to a cubic growth of the number of network parameters. To tackle this issue, in ComboGAN \cite{anoosheh2018combogan}, domain-specific encoders and decoders are combined to allow translation between any pair of domains. This approach leads to a linear growth of the number of parameters. To further reduce complexity, Choi~\etal~proposed StarGAN \cite{choi2018stargan} where a single generator can map images between any pair of domain leading to constant network complexity. In Table \ref{table:teaser}, we report the network complexity of these two approaches in terms of number of parameters and floating-point operation at inference time. While both methods have the same computational cost, ComboGAN has a prohibitively large number of parameters when the number of domain is high and StarGAN has a constant number of parameters. However, assuming a constant number of parameter may be problematic when the number of domains is high.

In addition, in this paper, we address the problem of multi-domain translation in the scenario of limited computation resources. In this case, only small architectures with few convolution filters can be employed. In this challenging scenario, existing approaches, such as the StarGAN, fail in generating realistic images since a single small architecture cannot handle the visual diversity of different domains. Therefore, in this paper, we propose a novel generator architecture named \textit{Ada$^2$Net}. This architecture is adaptive to two extents. First, the network topology is adapted on the fly depending on the input. In this way, the number of parameters can be adjusted while keeping a constant computational complexity in terms of floating-point operations (See Tab.~\ref{table:teaser}). By increasing the number of parameters, our model can handle a high number of visual domains. Second, we adopt the adaptive instance normalization layer to adapt the parameters of normalization layer parameters to the image style.

\begin{table}[t]
\centering
  \caption{Complexity comparison. We consider a common backbone encoder-decoder with C parameters and F floating point operations. In \textit{Ada$^2$Net}, we can adjust the parameter K depending on number of domains or memory constraints while keeping a constant computational cost. }
\resizebox{0.9\columnwidth}{!}{%
\begin{tabular}{ccc}
	\toprule
	& \textbf{Nbr. Params.} & \bf FLOPs \\
	\midrule
        ComboGAN \cite{anoosheh2018combogan} & 	CD & F \\
    StarGAN \cite{choi2018stargan} & 	C & F\\
    \textit{Ada$^2$Net} (Proposed) & CK	& $\approx$F \\

	\bottomrule
\end{tabular}
}
\label{table:teaser}
\end{table}

In short, our contributions are the following: (i) We propose an encoder-decoder network with an architecture that is dynamically chosen at inference time. To the best of our knowledge, this work is the first that employs dynamic network architectures for generation tasks. (ii) We impose a constant computation complexity by selecting specific sub-networks at inference time. Our framework is trained end-to-end via the use of Gumbel-Softmax for sub-network selection. (iii) Evaluation is performed on two widely used datasets of facial and painting images and shows that our method outperforms state-of-the-art approaches. Our code is available at https://github.com/thephucnguyen/Ada2Net.

\section{Related Work}
\label{sec:related_work}
Image-to-image translation consists in learning a mapping between two different domains. In the seminal work of Isola \etal~\cite{Isola_2017_CVPR}, a neural network is trained using paired data from the two domains. In many applications, paired data is not available or too expensive. To tackle this limitation, cycle consistency is used in CycleGAN \cite{zhu2017unpaired} to learn translation from unpaired images. As an alternative, geometry-consistency \cite{fu2019geometry,siarohin2019first,siarohin2019animating}~ can be imposed to obtain better translation results. Other recent works improve image quality by learning style-content disentangled representations \cite{Huang_2018_ECCV,DRIT} or including object-specific prior-knowledge \cite{Geng20193DGF,defGanPami,Tomei_2019_CVPR}.  

These works can be extended to the multiple domain setting. In this task, assuming an input image, we generate an image with the appearance of a specified target domain. To tackle this problem, ComboGAN \cite{anoosheh2018combogan} combines multiple encoder-decoder that are trained using cycle consistency principle introduced in CycleGAN\cite{zhu2017unpaired}. This solution leads to a prohibitively large architecture that cannot be used in our limited resources scenario. Conversely, a single encoder-decoder is trained in StarGAN \cite{choi2018stargan} to map any pair of domains. However, StarGAN generates low-quality images when the network architecture is reduced to satisfy our low resource constraints. In this work, we argue that a single architecture, especially small, can hardly handle a large number of domains since it requires learning a specific input image representation for every domain. Consequently, we introduce an adaptive network that can adjust its graph at inference time. Thus, we augment the network capacity without its computation cost at inference time.

In the context of image generation, adaptive instance normalization \cite{huang2017arbitrary} recently has become a popular tool. This layer can be seen as an adaptive layer since its parameters are not learned but predicted by another layer at inference time. Nevertheless, the network topology remains fixed. In parallel, several works proposed neural architectures with adaptive inference graphs\cite{Veit_2018_ECCV,Abati2020ConditionalCG} in order to reduce computation cost at inference time. In these works, network architecture is predicted on the fly with the aim of avoiding unnecessary operations. The network is trained end-to-end using Gumbel-softmax \cite{jang2017categorical} to obtain a differentiable approximation of the argmax function. In this work, we propose these two types of adaptive mechanisms. To the best of our knowledge, this work is the first attempt to use an adaptive network with a topology inferred on the fly in the context of generation tasks.

\begin{figure*}[t]
	\begin{center}
		\includegraphics[width=0.99\linewidth]{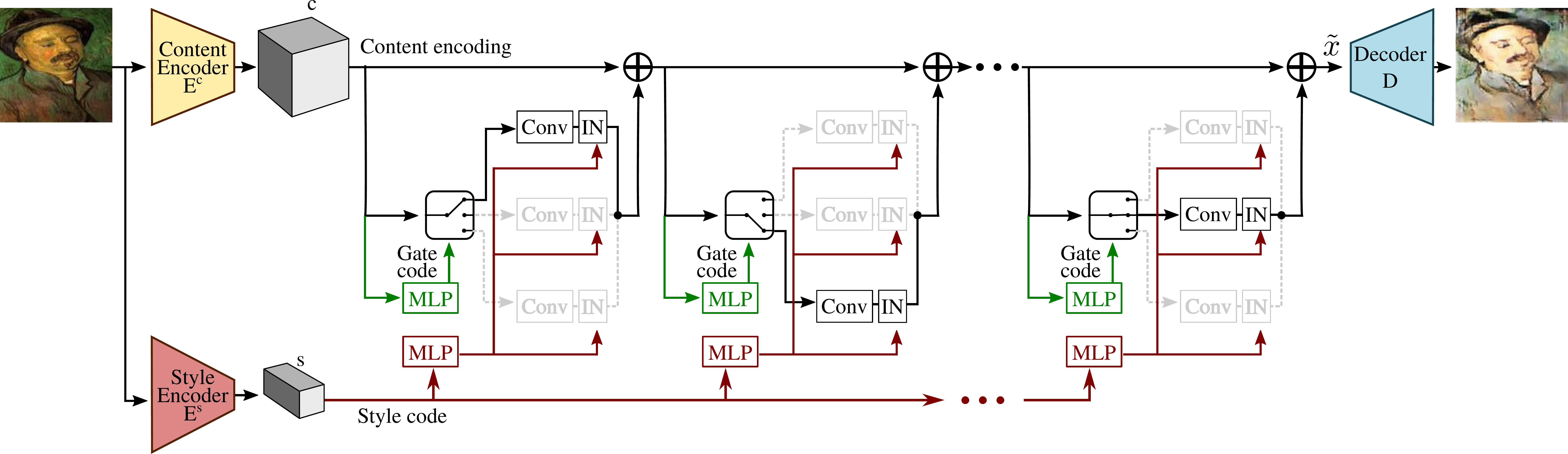}
	\end{center}
	\caption{\textit{Ada$^2$Net}: Two encoders extract content and style representations of the input. These two encodings are fed to a sequence of residual blocks with an adaptive graph. The network topology is predicted on the fly. Adaptive inference Normalization layers are used in each elementary block to obtain better style-content disentanglement. Finally, the resulting tensor is then provided to a small decoder network that generates the output image.}
	\vspace{0.25cm}
	\label{fig:mask}
\end{figure*}

\section{Ada$^2$Net }
\label{sec:method}
In this section, we detail the proposed architecture with an adaptive inference graph for multi-domain image-to-image translation. First, we describe the overall architecture. Then, we detail our gating mechanism for adaptive inference. Finally, we provide details concerning the training procedure.

\subsection{Overall architecture}\label{sec:overall}
Building on StarGAN framework \cite{choi2018stargan}, we perform image-to-image translation for multiple domains using only a single network $G$. Considering an input image $x$ from a source domain, the generator network translates $x$ into the domain $d$ as follows $x_d=G(x, d)$. The target domain label $d$ is provided to the generator network using one-hot encoding.

In this work, we further assume that each image $x$ contains content and style information. Following \cite{Huang_2018_ECCV}, we employ two encoder networks $E^c$ and $E^s$ that respectively extract a feature representation $c=E^c(x,d) \in \mathbb{R}^{H,W,C}$ describing content and a style code $s=E^s(x,d) \in \mathbb{R}^{S} $. Note that, in our low resource scenario, we use very small encoder networks.
In our multi-domain translation problem, this choice is motivated by the idea that content lies in a shared latent space between different domains while
style is specific to each domain. In other words, the content code $c$ encodes the shared information among all domains that should be preserved during translation, while the specific style code represents remaining variations from different domains that are not contained in the input image. 

The input image representations $c$ and $s$ are combined using a sequence of adaptive style-based Residual blocks described in the following section. In this way, we obtain a tensor $\tilde{x}=R(c,d)$ where $R$ denotes the sequence of adaptive style-based Residual blocks. This tensor $\tilde{x}$ is then input to a 2-layer decoder network $D$ that outputs the generated image.

Importantly, we disentangle content and style information thanks to the way the style code is used. Following recent style-based architectures \cite{huang2017arbitrary,Huang_2018_ECCV}, our style code is used only in the instance normalization layers included in the residual blocks. In instance normalization, the same affine transformation is applied at every location. Therefore, the style encoder extracts information that is not related to location and structure. Conversely, the content encoder network encodes what cannot be reconstructed from the style code.

\subsection{Adaptive style-based residual blocks}
In this section, we detail how we combine the content and style representation in order to obtain a single tensor. In place of commonly used residual blocks \cite{zhu2017unpaired,wang2018pix2pixHD}, we propose a novel adaptive style-based residual block. The motivation for adaptive residual blocks is that a small fixed architecture can hardly handle the visual diversity of different domains. 

A standard Residual network \cite{he2016deep} uses a "identity shortcut connection" that skips layers. Formally, assuming an input tensor $x_{l-1}\in \mathbb{R}^{H,W,C}$, the output of a standard residual block can be written as follows: 

\begin{equation}\label{eq:eq1.2}
    x_l = x_{l-1} + F_l(x_{l-1})
\end{equation}

where $F_l$ is a two-layer convolutional layer with ReLu.
In order to augment the capacity of the network, we propose to replace $F_l$ by  $K$ different residual functions $(F_l^1\hdots F_l^K)$. Nevertheless, in order to preserve a constant computational cost, only a single residual function is employed. Following \cite{Veit_2018_ECCV}, the layer input $x_{l-1}$ is fed to a small multi-layer perceptron $\Gamma_l$ that predicts which function should be used. More precisely, we assume that $\Gamma_l$ ends by a layer of dimension $K$ with softmax activation:  $\Gamma_l=(\gamma_l^1\hdots \gamma_l^K)$. Then, the output of our adaptive residual block is given by:

\begin{align}
  &x_l = x_{l-1} + F_l^{k^{*}}(x_{l-1}) \label{eq:adaREs}\\
  &\text{with } k^{*}= \underset{k}{\mathrm{argmax}}(\gamma_l^k(x_{l-1}))\nonumber
\end{align}

The intuition here is that each function $F_l^{k}$ can be specialized in a sub-region of the feature space. Therefore, we obtain a neural network with a larger capacity with a marginal computational over-head. Indeed, the total number of parameters of each block is multiplied by $K$ (ignoring the parameters of $\Gamma_l$). Concerning computation, the additional computation cost lies only in the gate network $\Gamma_l$. To reduce its cost, we use a global averaging pooling layer as the first layer of $\Gamma_l$. In this way, we obtain a computational over-head negligible with respect to the cost of $ F_l^{k^{*}}$.

The main difficulty in this approach is that the argmax function used in Eq.\eqref{eq:adaREs} needs to be implemented in a differentiable manner in order to allow end-to-end training of the whole network. Standard softmax cannot be used since it would imply that all the $F_l^{k}$ are computed leading to a high computational cost. In addition, replacing softmax by argmax at test time would introduce differences between training and test that may affect the results. To tackle this issue, we use the Gumbel-Max trick and its softmax relaxation proposed in \cite{jang2017categorical}. The key idea of this approach is that by sampling noise according to $K$ Gumbel distributions, we can obtain samples that follow the distribution corresponding to the probabilities provided by a softmax activation. Based on this idea, during training, in the forward pass, we employ discrete samples, while during the backward process, the gradient of the softmax is propagated. At test time, the argmax function is employed on the softmax predictions. For more details concerning the Gumbel-Max trick, please refer to \cite{jang2017categorical}.

Our adaptive residual blocks, as described so far, do not use any style information. Following recent trends in image generation, the style code is used to parametrize the instance normalization layer of each residual function $F_l$. This approach is usually referred to as adaptive instance normalization~\cite{huang2017arbitrary}. We employ a small network to predict the shift and scale parameter of each instance normalization layer. Formally, for a given layer input $x$, the output of the adaptive instance normalization layer can be written by:

\begin{equation}
  In(x,s)= \alpha(s)\Big(\frac{x-\mu(x)}{\sigma(x)}\Big) + \beta(s) \label{eq:IN}
\end{equation}
where $\alpha$ and $\beta$ denote the two heads of a small multi-layer perceptron.



\subsection{Training procedure}
\label{sec:training}
Inspired by \cite{choi2018stargan} and \cite{Huang_2018_ECCV}, our network is trained in an end-to-end manner combining cycle reconstruction losses and style-content disentanglement losses.

\noindent\textbf{Multi-domain Cycle reconstruction losses.} We use the cycle consistency ensued from CycleGAN \cite{zhu2017unpaired}.
First, we consider a reconstruction loss. Assuming an image $x$ from a domain $d$ and a target domain $d'$, the cycle reconstruction loss can be written:

\begin{equation}\label{eq:rec1}
  \mathcal{L}_{rec}^{cyc}= \mathbb{E}_{x,d^{'}}[||x-G(G(x,d'),d)||_1]
\end{equation}

We also include a reconstruction loss that enforces that the network reconstructs the input image when the target domain corresponds to the input domain:

\begin{equation}\label{eq:rec2}
  \mathcal{L}_{rec}^{in}= \mathbb{E}_{x,d}[||x-G(x,d)||_1]
\end{equation}

Then, in order to generate realistic images, we introduce a discriminator network and employ the least square adversarial framework from \cite{mao2017least}. It leads to two adversarial losses $\mathcal{L}_{adv}^{disc}$ and $\mathcal{L}_{adv}^{gen}$ that are used to update the discriminator and generator respectively. 
Finally, we employ a domain confusion loss. The goal of this loss is to assess that the generated image looks similar to images of the target domain. More specifically, we use a classifier $C_{cls}$ that is trained to predict the domain $d$ of a real image $x$:

\begin{equation}\label{eq:clas1}
    \mathcal{L}^{dis}_{cls} = \mathbb{E}_{x,d}[-\log C_{cls}(d|x)]
\end{equation}

In contrast, this domain-classifier is used to train the generator network using the following loss:

\begin{equation}\label{eq:clas2}
    \mathcal{L}^{gen}_{cls} = \mathbb{E}_{x,d^{'}}[-\log C_{cls}(d^\prime|G(x,d'))]
\end{equation}

By optimizing this objective, the generator $G$ tries to generate images that are classified as the target domain $d'$ by the classifier. In this way, we enforce that the generated images look similar to the real images of the targeted domain.
Practically, the domain-classifier is implemented as an auxiliary head of the discriminator used to compute the adversarial losses previously described.

\noindent\textbf{Style-content disentanglement losses.}
In order to obtain better style-content disentanglement, these reconstruction losses are completed with the style and content reconstruction losses from in \cite{Huang_2018_ECCV}. 
More precisely, for a given style code $s$, if an image is generated from $s$ by the decoder (\ie~$D(R(c,s))$), then, encoding this image with the style encoder should return a style code similar to the initial style code $s$:

\begin{equation}\label{eq:rec3}
  \mathcal{L}_{rec}^{s}= \mathbb{E}_{c,s,d}[||E^s(D(R(c,s)),d)-s||_1]
\end{equation}

\noindent Similarly, content consistency is imposed as follows:

\begin{equation}\label{eq:rec4}
  \mathcal{L}_{rec}^{c}= \mathbb{E}_{c,s,d}[||E^c(D(R(c,s)),d)-c||_1]
\end{equation}
where $c=E^c(x)$ and s is sampled according to a unitary Gaussian prior $\mathcal{N}(0,\textbf{I})$. For further mathematical discussion about these style-content disentanglement losses, please refer to \cite{Huang_2018_ECCV}.

\noindent\textbf{Total losses.}
Finally, we obtain two total objective functions for the generator and the discriminator respectively that can be written as follows:

\begin{align}\label{eq:eq1.7}
  \mathcal{L}^{gen}_{total} =& \lambda_{x}(\mathcal{L}_{rec}^{cyc} + \mathcal{L}_{rec}^{in}) + \lambda_{s}\mathcal{L}^{s}_{rec} + \lambda_{c}\mathcal{L}^{c}_{rec} \notag \\
  &+\lambda_{cls}\mathcal{L}^{gen}_{cls}+ \mathcal{L}_{adv}^{gen}
\end{align}
\begin{equation}\label{eq:eq1.8}
\mathcal{L}^{dis}_{total} = \mathcal{L}_{adv}^{dis} + \lambda_{cls}L^{dis}_{cls}
\end{equation}

\section{Experiments}
\label{sec:exp}
\subsection{Implementation details}
\paragraph{Network Architecture}
Let us introduce some notations to describe the network architecture. Let {\verb CkSs-f } denote a $k \times k$ convolution block with stride {\verb s }  and {\verb f } filters. {RkSs-k} denotes residual blocks that with two $k \times k$ convolution blocks. {\verb AbRkS1-f } denotes an adaptive residual blocks {\verb b } branches with two $k \times k$ convolution blocks with {\verb f } filters. {\verb U-f } denotes a nearest-neighbor up-sampling by 2 layer followed by a $5 \times 5$ convolution with stride 1 and {\verb f } filters. {\verb GAP } denotes a global average pooling layer and {\verb FC-f } denotes a fully connected layer with {\verb f } filters. \par
Employing this notation, the generator architecture can be written:
\begin{itemize}
    \item Content encoder: {\verb C7S1-64 }, {\verb C4S2-128 }, {\verb C4S2-256 }, {\verb R3S1-256 }, {\verb R3S1-R256 }, {\verb R3S1-R256 }, {\verb R3S1-R256 }
    \item Style encoder: {\verb C7S1-64 }, {\verb C4S2-128 }, {\verb C4S2-256 }, {\verb C4S2-256 }, {\verb C4S2-256 }, {\verb GAP }, {\verb FC-50 }
    \item Decoder: {\verb A3R3S1-256 }, {\verb A3R3S1-256 }, {\verb A3R3S1-256 }, {\verb A3R3S1-256 }, {\verb U-128 }, {\verb U-64 }, {\verb C7S1-3 } 
\end{itemize}

We employ a discriminator with multi-scale architecture with 3 scales using Leaky Relu as in \cite{zhu2017unpaired}. We employ the following discriminator architecture: {\verb C4S2-64 }, {\verb C4S2-128 }, {\verb C4S2-256 }, {\verb C4S2-512 }, {\verb C1S1-1+c(8:scale)s1-14 } (with {\verb scale }= 1,2,3 for 3 different scales)

\paragraph{Training hyper-parameters}
We train our model for $20,000$ iterations using the Adam optimizer \cite{kingma2014adam} with $\beta_{1} = 0.5$, $\beta_{2} = 0.999$ starting at learning rate of $0.0001$ and decaying by half after each $50,000$ iterations. In all experiments, we use a batch size of 16 and set the loss weights to $\lambda_{x} = 10$, $\lambda_{c} = 1$, $\lambda_{s} = 1$ and $\lambda_{cls} = 1$. Note that, these weighting parameters are taken from \cite{Huang_2018_ECCV} and \cite{choi2018stargan}. Despite the high number of loss terms, our model does not require further hyper-parameter tuning to converge properly.

\subsection{Datasets}
In our experiments, we employ two datasets:\\
\noindent
\textbf{Painting-14.} The Painting-14 dataset \cite{anoosheh2018combogan} contains approximately 10,000 paintings in total from 14 different artists from Wikiart.org. We follow the protocol used in \cite{anoosheh2018combogan}. We crop a square in all the images at the center and then resize them to $128 \times 128$. We randomly select 700 images from all artists as test set and use all remaining images for training. 

\noindent
\textbf{CelebA dataset.} We use CelebFaces Attributes (CelebA) dataset \cite{liu2015faceattributes} which is a large scale face attributes data set with large diversities, large quantities, and rich annotations, including 202,599 face images from 10,177 identities with 40 binary attributes annotations per image. In our experiments, we obtain seven domains using the following attributes: hair color (black, blond, brown), gender (male or female), and age (young or old). The generated images have a single attribute transfer in either hair color (black,  blond,  brown), gender, or age. Concerning image pre-processing , we firstly crop the original $178 \times 218$ images at the center to the size of $178 \times 178$ before resizing them as $128 \times 128$ images. The test set contains 2000 randomly chosen images and all remaining images are used for training. 

\begin{figure*}[t]
	\begin{center}
		\includegraphics[width=0.99\linewidth]{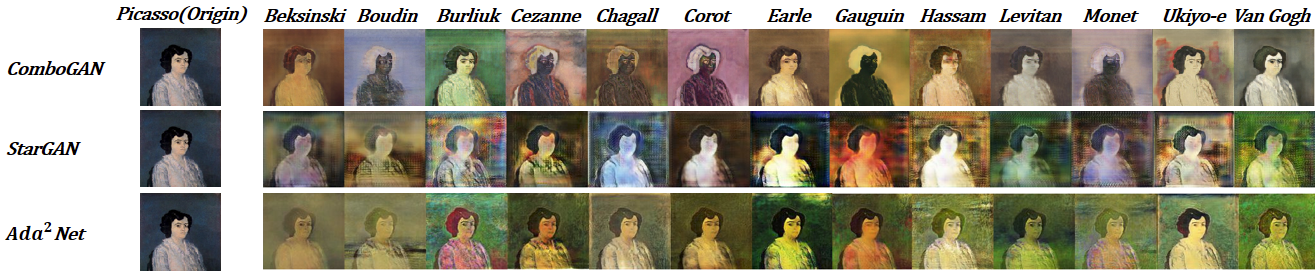}
	\end{center}
	\caption{Qualitative comparison with state-of-the-art on the Painting-14 dataset. The network receives as input the painting in the first columns and outputs the other images using the style indicated in each column.}
	\label{fig:result_1}
\end{figure*}

\subsection{Metrics}
For quantitative evaluation, we employ the Frechet Inception Distance (FID) score \cite{heusel2017gans} which is a popular metric for image generation tasks. The FID measures the similarity between the generated samples and the real ones in terms of statistics using a pre-trained Inception-v3 model\cite{szegedy2016rethinking}. The lower FID score is, the more similar two image distributions are. For evaluating the quality of images generated by generative adversarial networks, lower scores have been shown to correlate well with a higher quality of generated images \cite{heusel2017gans}. 
In order to calculate FID score, we firstly use the pre-trained Inception network to extract features from the 2048 dimensional activations of the Inception-v3 pool3 layer. After that, we assume Gaussian feature distributions for both real and generated images. The FID can be written: 
\begin{align}\label{eq:eq5.1}
  \mathcal{FID} =& \parallel m_r - m_g\parallel ^{2}_{2}+ \mathcal {T}_{r}({C}_{r}+{C}_{g}
   -2({C}_{r}{C}_{g})^{\frac{1}{2}})
\end{align}
where $\mathcal {T}_{r}$ sums up all the diagonal elements,  $m_r$ an ${C}_{r}$ denote the mean and covariance of the real image features while $m_g$ and ${C}_{g}$  the mean and covariance of generated image features.

\subsection{Comparison with State-of-the-Art}
For the qualitative comparison, we compare our \textit{Ada$^2$Net} with two literature methods: ComboGAN and StarGAN. Recent approaches such as DRIT~\cite{lee2018diverse}, MUNIT~\cite{Huang_2018_ECCV} or MSGAN~\cite{MSGAN} are not included in the comparison since they address the problem of translation between pairs of domains. On the contrary, in this work, we address the problem of multi-domain translation in the low resource regime. In this context, adapting these recent methods using multiple networks would not be fair.
Note that our architecture has a number of FLOPS and parameters similar to StarGAN at inference time. Figure \ref{fig:result_1} shows an example of style translation from a Picasso artwork to the remaining styles. We can observe that both ComboGAN and StarGAN fail to preserve the content of the input image. Most generated images by ComboGAN change the color of people to unrealistic colors. For instance, ComboGAN predicts very dark colors when translating in the styles of Boudin, Cezanne, Chagall, Corrot, Gauguin, and Monet. StarGAN affects the output image quality with many visible artifacts and abnormal bright regions. In addition, it also removes details such as the face of the character in the original painting.

By using the model with content and style extracted separately, our \textit{Ada$^2$Net} model produces images that are more diverse and realistic but also preserves better the content information of the input image. By using our model with content and style extracted separately, we obtain images that are more diverse and realistic and also that respect better the content information of the input image. The qualitative observations above are confirmed by our quantitative evaluations. \par

As shown in Table \ref{table:FID_sota}, the competitor methods obtain very high FID scores indicating that they generate unrealistic images with low visual quality. More precisely, the FID score of ComboGAN is quite high (154.3) due to unrealistic coloring as we can see in Fig.~\ref{fig:result_1}. StarGAN mitigates this issue but, the presence of artifacts, that alter the content, leads to a high FID score. Interestingly, \textit{Ada$^2$Net} approach obtains the lowest score for every single artist's style. It shows that our adaptive strategy succeeds in generating images of higher quality than state-of-the-art approaches.

\begin{table}[ht!]
\centering
\caption{Quantitative comparison with state-of-the-art: we compare the Frechet Inception Distance (FID) obtained with the different methods. The lower the better.}
\resizebox{\columnwidth}{!}{%
\begin{tabular}{cccc}
  \toprule
	\textbf{Artist} & \textbf{ComboGAN} & \textbf{StarGAN} & \textbf{\textit{Ada$^2$Net}}\\
 \midrule
	Beksinski & 179.05 & 146.05 & \bf 111.5\\
	Boudin & 160.3 & 148.54 & \bf 107.07\\
	Burliuk & 142.84 & 136.57 & \bf 105.16\\
	Cezanne & 146.44 & 129.98 & \bf 92.28\\
	Chagall & 142.66 & 109.96 & \bf 102.43\\
	Corot & 173.73 & 165.66 & \bf 106.37\\
	Earle & 178.68 & 172.73 & \bf 147.95\\
	Gauguin & 149.87 & 141.56 & \bf 98.51\\	
	Hassam & 140.98 & 141.24 & \bf 98.49\\
	Levitan & 151.83 & 186.92 & \bf 112.44\\
	Monet & 144.96 & 131.48 & \bf 78.81\\
	Picasso & 145.1 & 134.1 &  \bf123.69\\
	Ukiyo-e & 151.81 & 110.22 & \bf 99.97\\
	Van Gogh & 151.91 & 135.45 & \bf 97.25\\
\midrule
	\textbf{Average$\pm$Std} & 154.3$\pm$12.91 & 142.18$\pm$20.72 & \textbf{105.85$\pm$15.38}\\
\bottomrule
\end{tabular}
}
\label{table:FID_sota}
\end{table}

\subsection{Ablation study: adaptive graph}
\begin{figure*}[t]
	\begin{center}
		\includegraphics[width=0.99\linewidth]{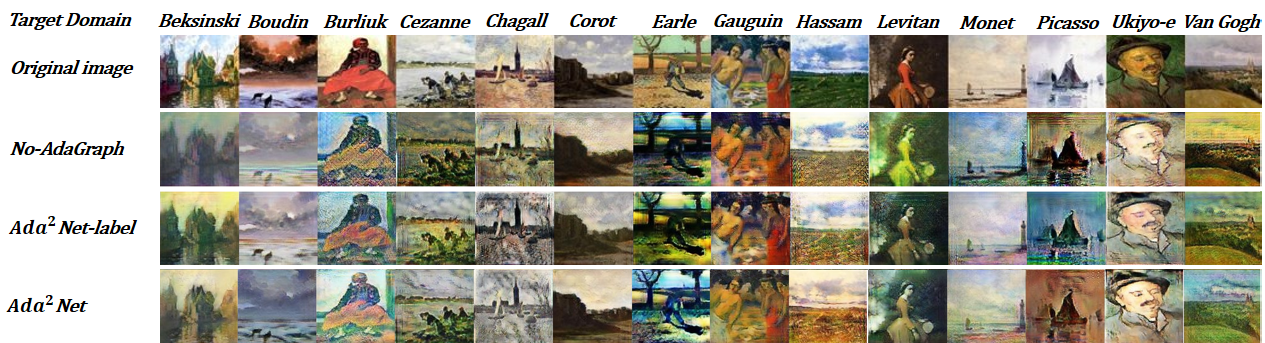}
	\end{center}
	\caption{Qualitative ablation study: impact of our adaptive network on style translation on the Painting-14 dataset. The network receives as input the painting in the first row and outputs the other images using the style indicated in each column.}
	\label{fig:result_2}
\end{figure*}
Finally, we empirically demonstrate that our architecture can increase the network performance without significant computational over-head and perform well on limited resources. 
We propose to compare three variants of our \textit{Ada$^2$Net} architecture:
\begin{itemize}
\item  \textbf{\textit{No-AdaGraph}}:  In this first baseline, we consider a standard architecture without adaptive graph. In this case, we use standard Residual blocks with adaptive inference Normalization layers.
\item \textbf{\textit{Ada$^2$Net-label}}: In this second model, we use a variant of our adaptive graph strategy where the gates are predicted by the labels. In other words, we use target labels as inputs of the gating units, instead of the image content. Therefore, the network topology depends only on the target domain label but not on the input image content.
\item \textbf{\textit{Ada$^2$Net}}: Finally, we evaluate our full \textit{Ada$^2$Net} model as described in Sec.\ref{sec:method} that uses the image content $c$ to control the network architecture. 
\end{itemize}
 In order to simulate the limited resources environment, we reduce the size of the model by reducing the number of filters in every convolution layer by 4 compared to the model used in the comparison with state of the art. This network size reduction is applied to the three baselines of the ablation study.
 
Qualitative comparison is reported in Figure \ref{fig:result_2}. First, we observe that the model without adaptive graph suffers from many artifacts. These defects are especially visible in the Burliuk, Cezanne, Hassam and Van Gogh styles. When we employ an adaptive network where gating is based on the domain labels, we obtain smaller artifacts. This improvement is especially clear in the domains Burliuk, Hassam, and Monet. When the graph is inferred from the content feature we further reduce the visual artifacts in the output images. A possible explanation for better performance of this content-based approach is that, in the case of the Painting-14 dataset, the images are really diverse both in terms of style and content. Indeed, the images depict a large diversity of scenes such as portrait, see-side landscape, country-side landscapes. In order to infer its graph, the network clusters the input space into sub-regions associated with a specific network graph. A possible interpretation of the better performance could be that clustering the input data according to the content is more efficient than clustering according to visual style. Still, this experiment shows that having a dynamic content-specific architecture improves the quality of the generated images.   

This qualitative analysis is completed by a quantitative analysis reported in Table \ref{table:FID_Ablation}. Indeed, the better qualitative performance of our adaptive approach is confirmed since the highest FID scores are obtained by \emph{No-AdaGraph}. The use of an adaptive graph improves the performance for almost every style and results in an average FID gain of $9.13$. Nevertheless, our full model that infers the network topology from the content vector leads to the best performance with an average FID of $134.35$. This quantitative analysis illustrates the potential of our adaptive strategy for multi-domain image-to-image translation.

\begin{table}[t]
\centering
  \caption{Ablation Study: we evaluate the impact on the performance in term of Frechet Inception Distance (FID) of the proposed adaptive graph model. The lower the better.}
\resizebox{\columnwidth}{!}{%
\begin{tabular}{cccc}
	\toprule
	\textbf{} & \textbf{\textit{No-AdaGraph}} & \textbf{\textit{Ada$^2$Net-label}} & \textbf{\textit{Ada$^2$Net}}\\
	\midrule
    Beksinski & 	168.63 & 151.79 &  \bf128.94\\
	Boudin & 	163.27 & 136.46 &  \bf127.84\\
	Burliuk &  	146.14 & 158.01 &  \bf146.17\\
	Cezanne & 	 	135.43 & 158.02 &  \bf 129.54\\
	Chagall &   	145.35  & 134.89 &  \bf128.51\\
	Corot &     	189.43 & 145.1 &  \bf126.47\\
	Earle &   	201.91 & 196.26 &  \bf186.77\\
	Gauguin &   	164.79 & 145.57 &  \bf131.67\\	
	Hassam &  	146.69 & 131.7 &  \bf121.51\\
	Levitan &  	148.12 & 140.86 &  \bf128.94\\
	Monet &   	99.06 & 99.46 &  \bf90.16\\
	Picasso &  	188.54 & 177.33  &  \bf167.19\\
	Ukiyo-e &  141.36 & 129.32  &  \bf125.85\\
	Van Gogh &  143.75 & 149.9  &  \bf131.66\\
	\midrule
	\textbf{Average$\pm$Std} & $155.89\pm 25.30$& $146.76\pm21.92$ &   $\bf 134.35\pm21.35$\\
	\bottomrule
\end{tabular}
}
\label{table:FID_Ablation}
\end{table}

\begin{table*}[t]
\centering
  \caption{Ablation Study: we evaluate the impact of each loss term on the performance in term of Frechet Inception Distance (FID). The lower the better.}
\begin{tabular}{cccccc}
	\toprule
	\textit{Target Domain} & \textbf{\textit{w/o classifier losses}} & \textbf{\textit{w/o reconstruction losses}} & \textbf{\textit{w/o content loss}} & \textbf{\textit{w/o style loss}}& \textbf{\textit{Ada$^2$Net}}\\
	\midrule
    Black Hair & 33.46	& 27.72 & 24.10 & 24.94   &\bf23.51\\
	Blond Hair & 53.02	& 53.32 & 42.09 & 44.14  &\bf41.82 \\
	Brown Hair & 39.77 	& 34.51 & 28.09 & 27.50   &\bf27.47\\
	Gender & 50.33	& 49.61 & 42.06 & 43.36  &\bf41.73\\
	Age &   43.69	& 39.82 & 35.48  & 36.12 &\bf34.94 \\
	\midrule
	 \textbf{Average$\pm$Std} &$44.05\pm7.91$ & $40.99\pm10.55$ & $34.37\pm8.14$ & $35.21\pm8.83$ & $\bf33.89\pm8.28 $\\
	\bottomrule
\end{tabular}
\label{table:FID_loss_Ablation}
\end{table*}

\begin{figure}[h]
    \centering
        \includegraphics[width=0.99\linewidth]{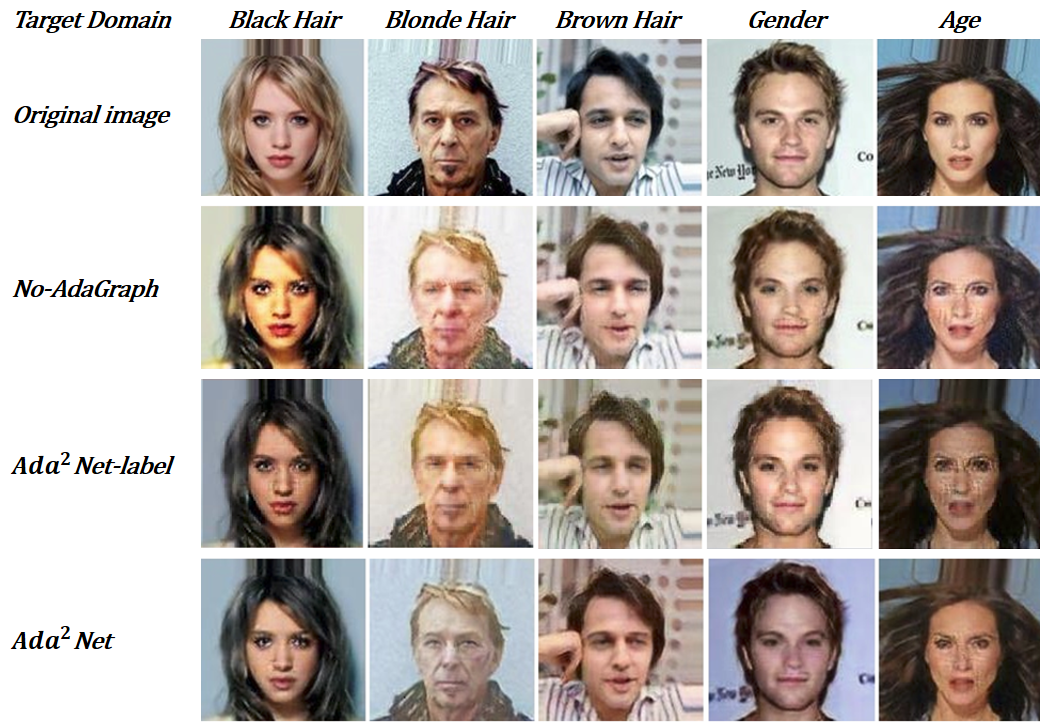}
    \caption{Qualitative ablation study on the CelebA dataset for attribute transfer.}
    \label{fig:result_3}
\end{figure}

We complete our ablation study by reporting additional results on CelebA dataset for the attribute transfer task. Qualitative results are reported in Figure \ref{fig:result_3}. The top row shows the original input images while the next three rows provide a qualitative comparison between the baselines \emph{No-AdaGraph}, \emph{Ada$^2$Net-$label$} and \emph{Ada$^2$Net}. The adaptive approach of \emph{Ada$^2$Net} shows a significant improvement since the generated images have fewer artifacts as well as sharper details. The attributes are also better translated. For instance, \emph{No-AdaGraph} does not succeed in transferring parts of the hair from blonde to black. Besides,  \emph{Ada$^2$Net-$label$} generates important artifacts in the \emph{Age} column. We note that the benefit of our adaptive graph model is smaller on this dataset than on the Painting-14 dataset. A possible explanation is that face images have a small diversity than paintings, and consequently, starGAN can handle all the face diversity pretty well.

\begin{figure}[h]
    \centering
        \includegraphics[width=0.99\linewidth]{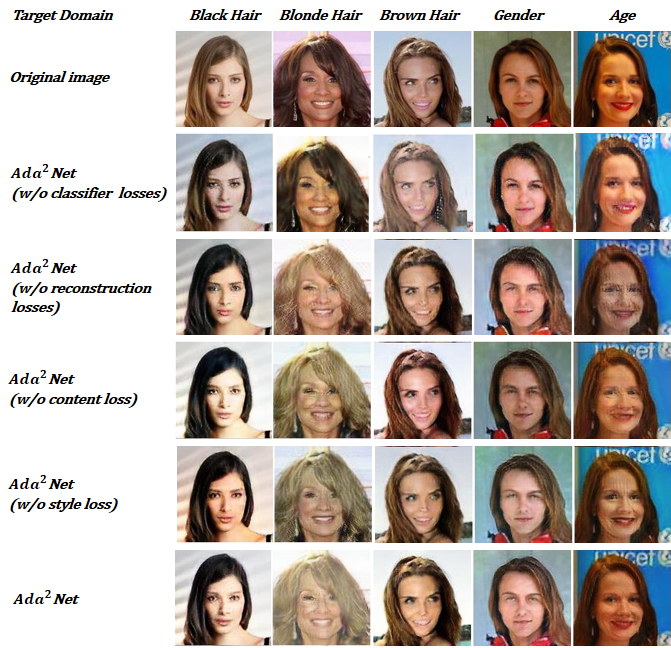}
    \caption{Qualitative ablation study on the CelebA dataset for the impact of each loss term.}
    \label{fig:ablationlosses}
\end{figure}

\subsection{Ablation study: training losses}
\label{sec:ablationLoss}
In this section, we perform an ablation study to show the impact of each loss term. We compare four baselines of \emph{Ada$^2$Net} where we remove the classifier losses (Eqs.\eqref{eq:clas1} and \eqref{eq:clas2}), the style loss (Eq.\eqref{eq:rec3}), the content  loss (Eq.\eqref{eq:rec4}) and the reconstruction losses (Eqs.\eqref{eq:rec1} and \eqref{eq:rec2}), respectively. Qualitative results are shown in Fig.~\ref{fig:ablationlosses}. First, we observe that, when we do not use the domain classifier (second row), the network generates realistic images but do not transfer the attribute. For example, in the first two columns, the women's hair are similar to the input images. When removing the style or the content loss, artifacts appear leading to poorer image quality. The artifacts are clearly visible in the \emph{Blond Hair} and \emph{Gender} columns. The artifacts are even stronger when the reconstruction losses are not used. These results are confirmed by the quantitative evaluation reported in Table~\ref{table:FID_loss_Ablation}. The model without domain-classifier reaches the highest FID showing that the generated images do not follow the distribution of the target domain. The average FID is also higher when the reconstruction losses are not used. Finally, the full \emph{Ada$^2$Net} model leads to the best FID score for every attribute. These consistent results show the positive impact of each loss term.



\section{Conclusions}
\label{sec:conclusions}
In this work, we presented \textit{Ada$^2$Net}, an adaptive method for the problem of multi-domain image-to-image translation. In our literature review, we observed that state-of-the-art methods require a large and deep model in order to handle the visual diversity of multiple domains. When computational resources are limited, increasing the network size may be problematic. Therefore, we introduced a novel network architecture with an adaptive graph structure. At inference time, the network estimates its own topology on the fly by selecting specific sub-network branches. The branch selection is implemented using Gumbel-Softmax to allow end-to-end training. This approach leads to an adjustable increase in number of parameters while preserving an almost constant computational cost. Our evaluation shows that \textit{Ada$^2$Net} can generate better images than literature methods. As future works, we plan to investigate this adaptive network for other structured problems such as instance segmentation or depth estimation \cite{ricci2018monocular}.






%
{
\bibliographystyle{IEEEtran}
\bibliography{bibfile}

\begin{thebibliography}{10}
\providecommand{\url}[1]{#1}
\csname url@samestyle\endcsname
\providecommand{\newblock}{\relax}
\providecommand{\bibinfo}[2]{#2}
\providecommand{\BIBentrySTDinterwordspacing}{\spaceskip=0pt\relax}
\providecommand{\BIBentryALTinterwordstretchfactor}{4}
\providecommand{\BIBentryALTinterwordspacing}{\spaceskip=\fontdimen2\font plus
\BIBentryALTinterwordstretchfactor\fontdimen3\font minus
  \fontdimen4\font\relax}
\providecommand{\BIBforeignlanguage}[2]{{%
\expandafter\ifx\csname l@#1\endcsname\relax
\typeout{** WARNING: IEEEtran.bst: No hyphenation pattern has been}%
\typeout{** loaded for the language `#1'. Using the pattern for}%
\typeout{** the default language instead.}%
\else
\language=\csname l@#1\endcsname
\fi
#2}}
\providecommand{\BIBdecl}{\relax}
\BIBdecl

\bibitem{Isola_2017_CVPR}
P.~Isola, J.-Y. Zhu, T.~Zhou, and A.~A. Efros, ``Image-to-image translation
  with conditional adversarial networks,'' in \emph{CVPR}, 2017.

\bibitem{zhu2017unpaired}
J.-Y. Zhu, T.~Park, P.~Isola, and A.~A. Efros, ``Unpaired image-to-image
  translation using cycle-consistent adversarial networks,'' in \emph{ICCV},
  2017.

\bibitem{Geng20193DGF}
Z.~Geng, C.~Cao, and S.~Tulyakov, ``3d guided fine-grained face manipulation,''
  \emph{CVPR}, 2019.

\bibitem{huang2017arbitrary}
X.~Huang and S.~Belongie, ``Arbitrary style transfer in real-time with adaptive
  instance normalization,'' in \emph{ICCV}, 2017.

\bibitem{couprie2018joint}
C.~Couprie, P.~Luc, and J.~Verbeek, ``Joint future semantic and instance
  segmentation prediction,'' in \emph{ECCV}, 2018.

\bibitem{pilzer2019refine}
A.~Pilzer, S.~Lathuiliere, N.~Sebe, and E.~Ricci, ``Refine and distill:
  Exploiting cycle-inconsistency and knowledge distillation for unsupervised
  monocular depth estimation,'' in \emph{CVPR}, 2019.

\bibitem{anoosheh2018combogan}
A.~Anoosheh, E.~Agustsson, R.~Timofte, and L.~Van~Gool, ``Combogan:
  Unrestrained scalability for image domain translation,'' in \emph{CVPR
  Workshops}, 2018.

\bibitem{choi2018stargan}
Y.~Choi, M.~Choi, M.~Kim, J.-W. Ha, S.~Kim, and J.~Choo, ``Stargan: Unified
  generative adversarial networks for multi-domain image-to-image
  translation,'' in \emph{CVPR}, 2018.

\bibitem{fu2019geometry}
H.~Fu, M.~Gong, C.~Wang, K.~Batmanghelich, K.~Zhang, and D.~Tao,
  ``Geometry-consistent generative adversarial networks for one-sided
  unsupervised domain mapping,'' in \emph{CVPR}, 2019.

\bibitem{siarohin2019first}
A.~Siarohin, S.~Lathuili{\`e}re, S.~Tulyakov, E.~Ricci, and N.~Sebe, ``First
  order motion model for image animation,'' in \emph{Advances in Neural
  Information Processing Systems}, 2019, pp. 7137--7147.

\bibitem{siarohin2019animating}
------, ``Animating arbitrary objects via deep motion transfer,'' in
  \emph{CVPR}, 2019.

\bibitem{Huang_2018_ECCV}
X.~Huang, M.-Y. Liu, S.~Belongie, and J.~Kautz, ``Multimodal unsupervised
  image-to-image translation,'' in \emph{ECCV}, 2018.

\bibitem{DRIT}
H.-Y. Lee, H.-Y. Tseng, J.-B. Huang, M.~K. Singh, and M.-H. Yang, ``Diverse
  image-to-image translation via disentangled representations,'' in
  \emph{ECCV}, 2018.

\bibitem{defGanPami}
A.~{Siarohin}, S.~{Lathuilière}, E.~{Sangineto}, and N.~{Sebe}, ``Appearance
  and pose-conditioned human image generation using deformable gans,''
  \emph{IEEE TPAMI}, 2019.

\bibitem{Tomei_2019_CVPR}
M.~Tomei, M.~Cornia, L.~Baraldi, and R.~Cucchiara, ``Art2real: Unfolding the
  reality of artworks via semantically-aware image-to-image translation,'' in
  \emph{CVPR}, 2019.

\bibitem{Veit_2018_ECCV}
A.~Veit and S.~Belongie, ``Convolutional networks with adaptive inference
  graphs,'' in \emph{ECCV}, 2018.

\bibitem{Abati2020ConditionalCG}
D.~Abati, J.~M. Tomczak, T.~Blankevoort, S.~Calderara, R.~Cucchiara, and B.~E.
  Bejnordi, ``Conditional channel gated networks for task-aware continual
  learning,'' in \emph{CVPR}, 2020.

\bibitem{jang2017categorical}
E.~Jang, S.~Gu, and B.~Poole, ``Categorical reparametrization with
  gumbel-softmax,'' in \emph{ICLR}, 2017.

\bibitem{wang2018pix2pixHD}
T.-C. Wang, M.-Y. Liu, J.-Y. Zhu, A.~Tao, J.~Kautz, and B.~Catanzaro,
  ``High-resolution image synthesis and semantic manipulation with conditional
  gans,'' in \emph{CVPR}, 2018.

\bibitem{he2016deep}
K.~He, X.~Zhang, S.~Ren, and J.~Sun, ``Deep residual learning for image
  recognition,'' in \emph{CVPR}, 2016.

\bibitem{mao2017least}
X.~Mao, Q.~Li, H.~Xie, R.~Y. Lau, Z.~Wang, and S.~Paul~Smolley, ``Least squares
  generative adversarial networks,'' in \emph{ICCV}, 2017.

\bibitem{kingma2014adam}
D.~Kingma and J.~Ba, ``Adam: A method for stochastic optimization,''
  \emph{ICLR}, 2015.

\bibitem{liu2015faceattributes}
Z.~Liu, P.~Luo, X.~Wang, and X.~Tang, ``Deep learning face attributes in the
  wild,'' in \emph{ICCV}, 2015.

\bibitem{heusel2017gans}
M.~Heusel, H.~Ramsauer, T.~Unterthiner, B.~Nessler, and S.~Hochreiter, ``Gans
  trained by a two time-scale update rule converge to a local nash
  equilibrium,'' in \emph{NeurIPS}, 2017.

\bibitem{szegedy2016rethinking}
C.~Szegedy, V.~Vanhoucke, S.~Ioffe, J.~Shlens, and Z.~Wojna, ``Rethinking the
  inception architecture for computer vision,'' in \emph{CVPR}, 2016.

\bibitem{lee2018diverse}
H.-Y. Lee, H.-Y. Tseng, J.-B. Huang, M.~Singh, and M.-H. Yang, ``Diverse
  image-to-image translation via disentangled representations,'' in
  \emph{ECCV}, 2018.

\bibitem{MSGAN}
Q.~Mao, H.-Y. Lee, H.-Y. Tseng, S.~Ma, and M.-H. Yang, ``Mode seeking
  generative adversarial networks for diverse image synthesis,'' in
  \emph{CVPR}, 2019.

\bibitem{ricci2018monocular}
D.~Xu, E.~Ricci, W.~Ouyang, X.~Wang, N.~Sebe \emph{et~al.}, ``Monocular depth
  estimation using multi-scale continuous crfs as sequential deep networks,''
  \emph{IEEE transactions on pattern analysis and machine intelligence},
  vol.~41, no.~6, pp. 1426--1440, 2018.

\end{thebibliography}
}

\end{document}